# *Shine*: A Deep Learning-Based Accessible Parking Management System

Dhiraj Neupane[a,1], Aashish Bhattarai[b], Sunil Aryal[a], Mohamed Reda Bouadjenek[a], Uk-Min Seok[b], Jongwon Seok[c,*]

[a]*School of Information Technology, Deakin University, Waurn Ponds Campus, Geelong, VIC 3216, Australia*
[b]*IP Camp [(주)아이피 캠프], Jinju-si, Gyeongsangnam-do 52818, South Korea*
[c]*Department of Information and Communication Engineering, Changwon National University, Changwon-si, Gyeongsangnam-do 51140, South Korea*

**Abstract**

The ongoing expansion of urban areas facilitated by advancements in science and technology has resulted in a considerable increase in the number of privately owned vehicles worldwide, including in South Korea. However, this gradual increment in the number of vehicles has inevitably led to parking-related issues, including the abuse of disabled parking spaces (hereafter referred to as accessible parking spaces) designated for individuals with disabilities. Traditional license plate recognition (LPR) systems have proven inefficient in addressing such a problem in real-time due to the high frame rate of surveillance cameras, the presence of natural and artificial noise, and variations in lighting and weather conditions that impede detection and recognition by these systems. With the growing concept of parking 4.0, many sensors, IoT and deep learning-based approaches have been applied to automatic LPR and parking management systems. Nonetheless, the studies show a need for a robust and efficient model for managing accessible parking spaces in South Korea. To address this, we have proposed a novel system called, *'Shine'* , which uses the deep learning-based object detection algorithm for detecting the vehicle, license plate, and disability badges (referred to as cards, badges, or access badges hereafter) and verifies the rights of the driver to use accessible parking spaces by coordinating with the central server. Our model, which achieves a mean average precision of 92.16%, is expected to address the issue of accessible parking space abuse and contributes significantly towards efficient and effective parking management in urban environments.

---

*Corresponding author.
 *Email addresses:* d.neupane@deakin.edu.au; dhirajneupane1717@gmail.com (Dhiraj Neupane),
apandey9860@gmail.com (Aashish Bhattarai), sunil.aryal@deakin.edu.au (Sunil Aryal),
reda.bouadjenek@deakin.edu.au (Mohamed Reda Bouadjenek), dolphung5524@naver.com (Uk-Min Seok),
jwseok@changwon.ac.kr (Jongwon Seok )

---

**1. Introduction**

Urban cities are being expanded nowadays as the population and business activities are escalating rapidly due to the enhancement in science and technology. As a result, the number of vehicles, primarily privately owned cars, is ascending dramatically every day. There are about 1.474 billion cars worldwide as of 2023, according to the Hedges & Company [1]. As the number of vehicles continues to escalate, the issue of parking has become a significant challenge in both developing and developed countries, and it has been among the most discussed topics by the general public and elites Ibrahim (2017). The Republic of Korea, commonly known as South Korea, has accumulated a registered car count exceeding 25 million as of 2023[2], with a notable increase in numbers being observed regularly. The rise in car ownership is primarily attributed to the numerous perceived advantages of personal vehicle possession. However, the simultaneous rise in vehicle volume has led to a surging weight of responsibility upon the country's traffic management and vehicle supervision systems. The inadequate parking infrastructure directly exacerbates issues such as traffic congestion, road accidents, air pollution, and the persistent shortage of available parking slots KC & Kang (2019). The ever-increasing demand for parking spaces has resulted in a corresponding rise in parking violations, which includes encroaching upon the rights of other drivers and abusing parking spaces allocated for individuals with disabilities Gining et al. (2018). The lack of proper management of accessible parking spaces leads to such illegal activities. As per the report by the Statista[3], more than 2.6 million individuals, which is about 5% of the total population of South Korea, are registered as persons with disabilities. A substantial proportion of vehicles are owned or operated by individuals with disabilities. Thus, a proper and practical approach is needed to solve the problem of accessible parking violations.

To comprehensively understand the proposed system for managing accessible parking spaces in South Korea, it is essential to introduce the characteristics of South Korean license plates (LP). LP is the primarily used and most effective way to identify the vehicle for both traditional and advanced automatic license plate recognition (ALPR) systems. Effective from September 2019, the South Korean government adopted a new LP consisting of seven digits, a Korean (Hangul) character, and a hologram. Table 1 provides detailed information regarding the various kinds of existing private vehicle number plates used in South Korea Usmankhujaev et al. (2020). The # character denotes the digits, **H** represents the Hangul character, and **R** denotes the regional character (city or province) used in the LP. The Korean LP format—*types a* to *c*—are new and existing formats, while *types d* and *e* are old; however, some vehicles still use these LP formats. Additional information on South Korean LP formats can be found in Wang et al. (2021b). It is crucial to note that fixed Hangul characters, **H**, and regional character, **R**, have been used in LP of private vehicles in South Korea. Table 1 shows additional information regarding the characters used in the Korean LP.

---

[1]https://hedgescompany.com/blog/2021/06/how-many-cars-are-there-in-the-world
[2]https://www.statista.com/statistics/1095105/south-korea-number-of-vehicles-registered/
[3]https://www.statista.com/statistics/1250146/south-korea-total-number-of-registered-disabled-persons/

Korean government provides a particular card, called a disability or access badge, to people with disability. These badges are yellow, white, and brown coloured. Each badge contains an accessible logo and the 4-digits card number linked to the LP number of a badge holder's vehicle. Usually, the card number is the last four digits of the LP. The white card is provided to the person with a disability, and the yellow card is to the guardians of the disabled person. The institution related to the welfare of people with disabilities is also provided with a yellow card [4]. The brown cards are provided to a disabled person of national merit. These cards need to be put in the front glass of the vehicle in such a way that they can be easily and properly visualised. Figure 1(a) shows the car and card position picture, and figure 1(b) shows the card type: brown (*ucard*), yellow, and white cards (from top to bottom).

---

[4]https://www.bsgangseo.go.kr/welfare/contents.do?mId=0701010400

| Type | Year | Number Format | Dimension (mm) | Example | Remarks |
|---|---|---|---|---|---|
| a | 2019- | ###H#### | 520×110- (European Sized) | 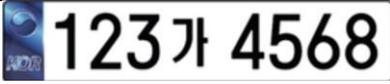 | Hologram, White Plate |
| b | 2017- | ##H#### | 520×110- (European Sized) | 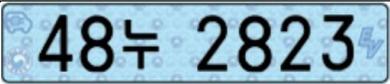 | Electric Car, Blue Plate |
| c | 2006-2019 | ##H#### | 335×155 (North American Sized) | 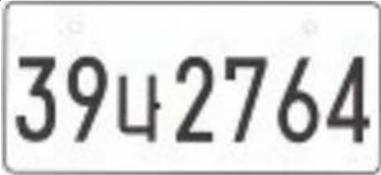 | White Plate |
|   |   |   | 520×110 | 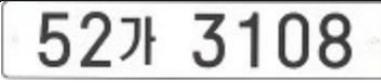 |   |
| d | 2004-2006 | ##H #### | 335×170 | 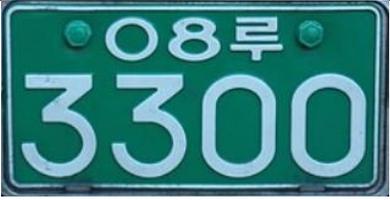 | Green Plate, Old, Exist some |
| e | 1973-2003 | R## H#### | 335×170 | 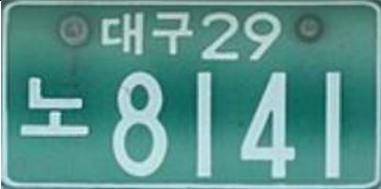 | Green Plate, Very Old, exist only few |

Table 1: An overview of South Korean private vehicle LP formats. This table outlines the various LP formats used in South Korea, including the format type, year of introduction, number format, dimensions (in mm), example images, and remarks to highlight specific details of each format.

|  |  | **Remarks** |
|---|---|---|
| **Hangul Characters (H)** | 가, 나, 다, 라, 마, 거, 너, 더, 러, 머, 버, 서, 어, 저, 고, 노, 도, 로, 모, 보, 소, 오, 조, 구, 누, 두, 루, 무, 부, 수, 우, 주 | Private Cars |
|  | 허,하,호 | Rental Cars |
| **Regional Characters (R)** | 강원, 경기, 경남, 경북, 광주, 대구, 대전, 부산, 서울, 세종, 울산, 인천,전남, 전북, 제주, 충남, 충북 | Province or City |

Table 2: Hangul and regional characters used in South Korean LP.

Different approaches have been practised around the globe for license plate recognition and parking management systems. However, only a few works can be found on LPR for disability parking management systems. The traditional approach of monitoring by the authorities manually used to be the dominating one a couple of years ago. It is still practised in some places to date. However, it is troublesome and not so effective. This approach solely depends on the visual inspections by parking attendants, which is costly and ineffective because of the challenge of detecting unauthorised users ultimately Gining et al. (2018). Similarly, the use of the internet of things (IoT) and sensor-based techniques is also found to be ubiquitous. Studies Gining et al. (2018); Fikri & Hwang (2019); Gining et al. (2021) and use of IoT technology in North of Spain [5] are some of the works based on IoT and sensors. The use of such IoT and sensor-based methods come up with some disadvantages: uneconomical because the sensors are expensive, need higher installation cost, require regular maintenance, and are prone to a security threat and connection breakdown Kannan et al. (2022), need experts for operation, and so on.

LPR is a complex and challenging task. It is mainly because the LPR systems must deal with various conditions: low-light, low-resolution, sunny, shiny, rainy, and foggy environments. The sensors or IoT-based systems need to provide correct data in varying and harsh weather, which is challenging and expensive. Due to the adverse effects of the environment, the LP images obtained are often of low quality: low light, low contrast, low brightness, and blur, hampering the efficiency of image-based systems like OpenCV or deep learning (DL) methods. Also, different countries, sometimes different provinces in the same country as well, have their LP with different fonts, styles, dimensions, and formats Usmankhujaev et al. (2020); Wang et al. (2021b), which restricts the availability of large datasets needed for image-based deep learning methods. The change and updates in the LP format from the policy level also somewhat affect this field.

---

[5]https://www.libelium.com/libeliumworld/success-stories/iot-technology-to-monitor-parking-for-disabled-citizens-in-the-north-of-spain

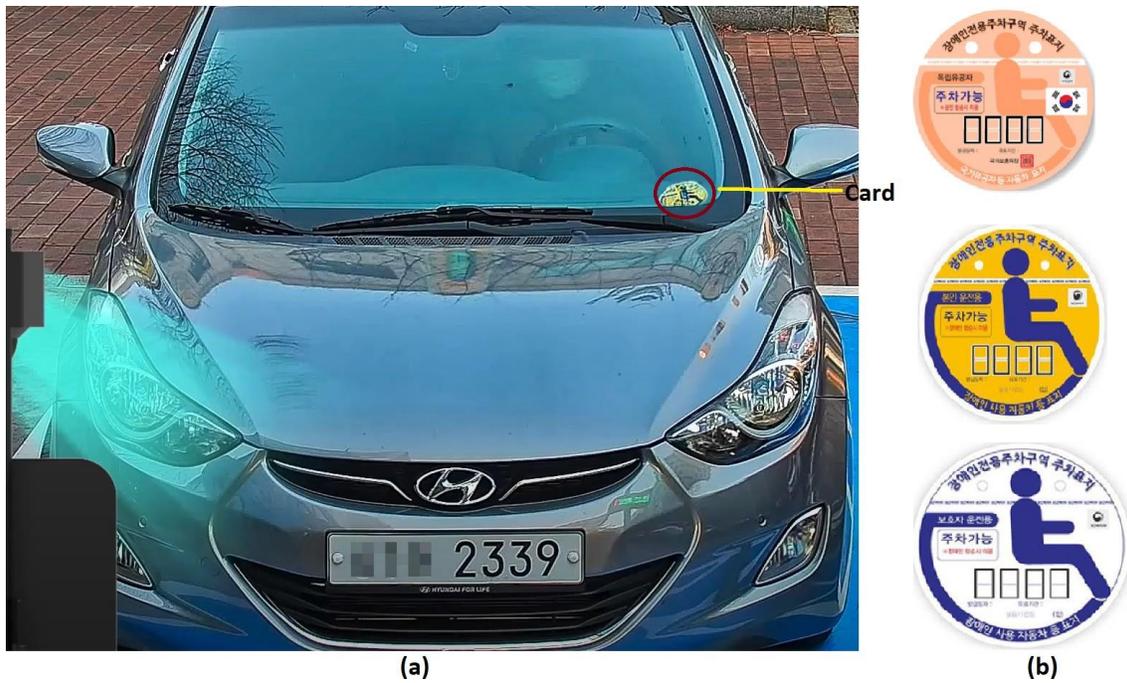

Figure 1: (a) A clear illustration of a car with a disability card on its front windshield, demonstrating the proper placement of the card. (b) A collection of disability access cards is used in South Korea, with brown (*ucard*), yellow, and white cards arranged from top to bottom.

To overcome the problems mentioned above, we have used a DL-based object detection (OD) algorithm, 'You Only Look Once' version 7 (YOLOv7), for the accessible parking management system. We have named our system *'Shine'* (Korean: 샤인) for the sake of clear communication and convenience during the research and development phase. Detailed information about this system is presented in the later section. Deep learning techniques are the current buzzwords with breakthrough results in computer vision, object detection, semantic segmentation, natural language processing, medical sciences, machinery fault detection, etc. These algorithms have attracted the attention of many scientists because of their efficient workability on large amounts of data, consistent problem-solving approach, and pleasing results Neupane & Seok (2020b), 2021). Also, in parking management systems, DL methods are seen to provide efficient results. Thus, we have used a transfer learning (TL)-based OD model for managing the accessible parking area, mitigating the violation of the parking space by a normal user. In this research, we have trained a YOLOv7 base model using images consisting of vehicles in different environments to increase the accuracy of our models in real world scenarios. The model is used for detecting the vehicles, varieties of LPs, the LP number, access badge, and card number. We have also compared the results obtained by using YOLOv7 with those obtained from previous YOLO versions, YOLOv4 Darknet and YOLOv5s, and with Resnet-50 backboned faster R-CNN model.

## 2. 2. Related Theory and Works

*2.1. Related Theory*

*2.1.1. Transfer Learning (TL)*

Typical machine learning (ML) systems deal with isolated tasks. Transfer learning attempts to change this situation by developing ways to transfer knowledge learned on one or more source tasks to improve learning on related target tasks. TL is machine learning that adds additional sources of information outside the standard training data. ML and DL algorithms are effective when the amount of data is enormous. However, in practice, large amounts of data are not always available conveniently. Therefore, a transfer learning approach is proposed to overcome this problem. It solves the problem of insufficient training data. Figure 2 shows the overview of TL approach. In this algorithm, information flows only in one direction, from the source task to the target task Neupane & Seok (2020a,c). Pre-training is a crucial component of TL and can be done through standard supervised, weakly supervised, and unsupervised learning to help the model develop a robust feature extractor. Note that pre-training for the target task should be done on a broader scope to ensure that "negative transfer" does not occur Jabed & Shamsuzzaman (2022). Here,"negative transfer" refers to the situation where pre-training on a source task hinders or negatively affects the model's performance on the target task. The careful selection of the pre-training task and its relevance to the target task must be ensured to prevent negative transfer. We have used a pre-trained YOLOv7 model after configuring it as per our need in this research.

*2.1.2. Object Detection (OD)*

Classification involves determining the category or identifying the specific label associated with an object. Localization, on the other hand, involves pinpointing the precise position of an object within an image, typically achieved by outlining bounding boxes around the object. Object detection encompasses both classifying and localizing one or more objects within an image, representing a fundamental challenge within the field of computer vision. The pipeline of typical object recognition models can be divided into three main stages: informative region selection, feature extraction, and classification Zhao et al. (2019). Thus, OD determines the class to which an object belongs, and the object's position is estimated by outputting a bounding box around the object. Detecting a single instance of a class in an image is called single-class OD, whereas detecting multiple classes that appear in an image is called multi-class object detection Pathak et al. (2018).

Real-time OD is a crucial component of computer vision systems. For example, autonomous driving Feng et al. (2020), multi-object tracking Zhang et al. (2021), medical image analysis Liang et al. (2022), robotics Karaoguz & Jensfelt (2019), etc. With the development of powerful graphics processing units (GPU), tensor processing units (TPU), and neural processing units (NPU), there has been a breakthrough achievement in the field of computer vision. The researchers are currently focusing on developing efficient real-time OD algorithms. The most popular real-time object detectors are region-based convolutional neural network (RCNN) series Girshick et al. (2014); Girshick (2015); Ren et al. (2015); He et al. (2017), YOLO series Redmon et al. (2016); Redmon & Farhadi (2017, 2018); Bochkovskiy et al. (2020); Jocher et al. (2022); Wang et al. (2023), MobileNets series Howard et al. (2017); Sandler et al. (2018); Howard et al. (2019), single shot detector (SSD) Liu et al. (2016), YOLOX Ge et al. (2021), YOLOR Wang et al. (2021a), MCUNet Lin et al. (2020), NanoDet Li & Zhai (2022), and so on. Currently, DL-based object recognition algorithms usually fall into two categories: the first is two-stage algorithms based on R-CNN and TridenNet Li et al. (2019). Existing problems with these two-stage algorithms include low real-time performance, large model sizes, and poor recognition efficiency for small objects. The second is a single-stage algorithm based on SSD and YOLO, which provides high real-time performance when detecting objects at multiple scales. Among them, the YOLO series has been prevalent, and we see the updated version every year.

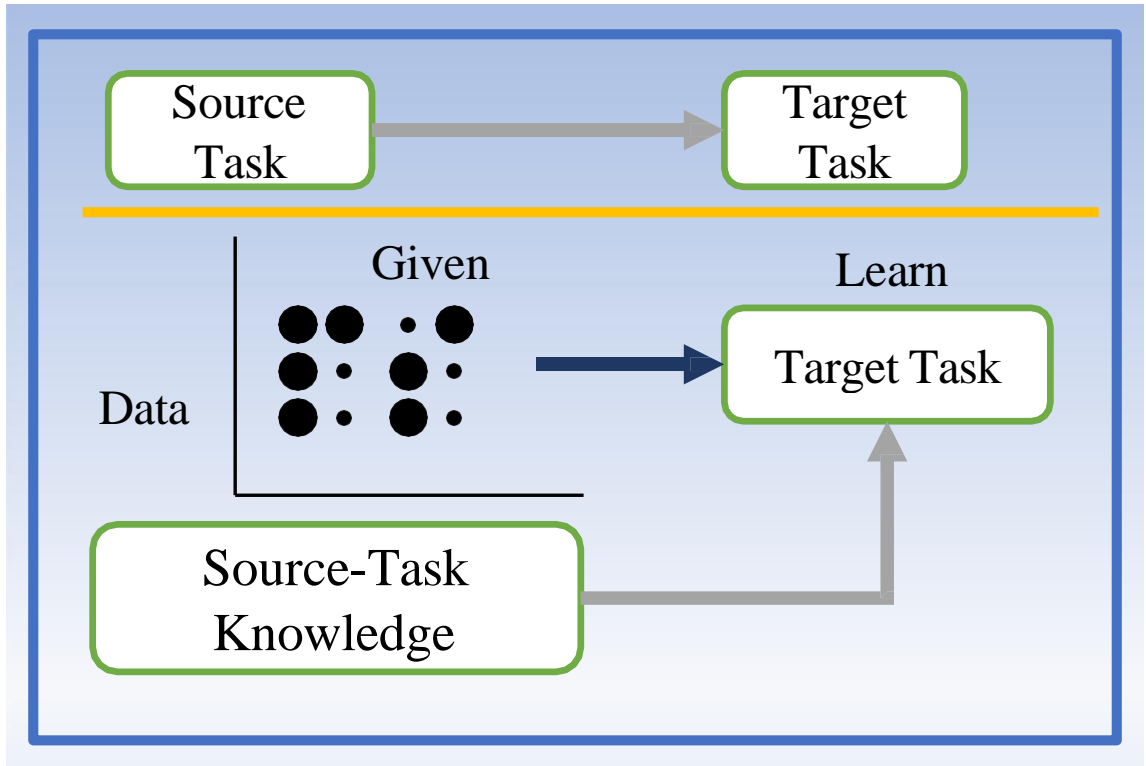

Figure 2: A visual representation of the transfer learning algorithm, showing the process of using pre-trained knowledge from a source model to improve the performance and efficiency of the target model.

You Only Look Once, commonly known as YOLO, is a widely adapted single-stage multiple OD algorithm. Joseph et al. Redmon et al. (2016) introduced the first version of YOLO in 2015. This is one of the most effective algorithms that has been a milestone for encompassing numerous innovative ideas for the computer vision research community. YOLO version 1 (YOLOv1) consists of 24 convolutions and two fully connected layers. It predicts multiple bounding boxes per grid cell. The bounding box with the highest intersection over union (IOU) with the ground truth is selected, known as non-maximum suppression. The core of the YOLO algorithm lies in its small model size, simple architecture, and fast computational speed. Bounding box positions and categories can be output directly through the neural network Jiang et al. (2022). With every version, the researchers have mitigated the shortcomings of the YOLO algorithm and made it the most popular and adapted OD module. The YOLO version used in this research is YOLOv7. Also, YOLOv4 and YOLOv5s are used for comparison purposes.

1. **YOLOv7:** The latest YOLOv7 version, released in July 2022, is claimed to be the most powerful, accurate and fastest single-stage real-time multiple OD algorithm till the writing of this article. This version is making waves in computer vision and machine learning. It requires less expensive hardware than other neural networks and can be trained much faster on smaller datasets without pre-trained weights. Therefore, YOLOv7 is expected to become the industry standard for object recognition soon, surpassing the earlier state-of-the-art in real-time applications. The YOLO framework consists of three main elements: neck, head, and backbone. The primary function of the backbone is to extract critical information from the image and send it to the head through the neck. In the neck, feature maps extracted by the backbone are compiled to create a feature pyramid. The output layer of the head with the final detection forms the last part of the structure. YOLOv7 is not limited to a single head. The main head is responsible for the final output, while the auxiliary head is responsible for supporting the learning of the intermediate layers. Again, to improve the learning of the deep network, a label assignment mechanism was introduced that considers both the network's predicted results and the ground truth and then assigns soft labels. While conventional label assignment relies on ground truth and generates hard labels based on predefined rules, reliable soft labels use computational and optimisation methods that consider the quality and distribution of the prediction results and ground truth. The latest YOLOv7 algorithm outperforms previous object recognition models and versions of YOLO in speed and accuracy (56.8% average precision). Figure 3 shows the performance comparison of YOLOv7 with other popular real-time object detectors Jabed & Shamsuzzaman (2022); Wang et al. (2023); Yang et al. (2022).

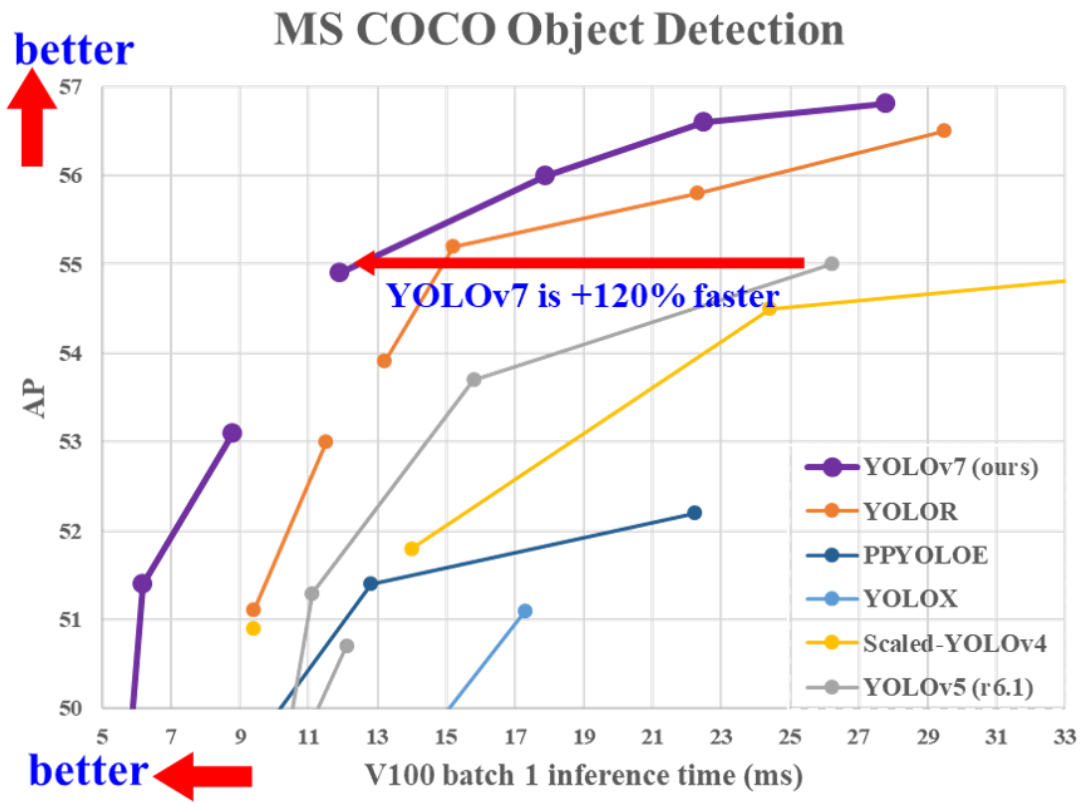

Figure 3: Performance comparison of YOLOv7 with other real-time object detection models, as presented in the original YOLOv7 paper Wang et al. (2023). This illustration highlights the relative performance of YOLOv7 compared to other models, providing valuable context for our study. Please note that the performance evaluation illustrated here is based on the results from the original research and not our own.

2. **YOLOv5:** YOLOv5 was released in May 2020 by Ultralytics. It is available in four models: s, m, l, and x, which differ in detection accuracy and performance. The letters 's', 'm', 'l', and 'x' denote 'small', 'medium', 'large', and 'extra-large'. YOLOv5s is the simplest version, with smaller weight files and faster detection speed. The mean average precision (mAP) of YOLOv5s is 55.6%. The significant upside of the YOLOv5 model is its smaller model size, i.e., about 27 megabytes. YOLOv5 gets simple support and deployment benefits from PyTorch as it was first carried out in the PyTorch framework. YOLOv5 is almost identical to YOLOv4 except for a few differences. YOLOv4 is published in the Darknet framework written in C, while YOLOv5 is based on the PyTorch framework. YOLOv4 uses a '.cfg' file for configuration, whereas YOLOv5 uses a '.yaml' file Jocher et al. (2022); Karthi et al. (2021).

3. **YOLOv4:** YOLOv4 is a real-time high-precision one-stage regression-based object recognition algorithm proposed in April 2020 that integrates features from YOLOv1, YOLOv2, YOLOv3, and others to achieve the current optimum in the trade-off between recognition speed and accuracy. Deeper and more complex networks are developed using dense blocks to achieve higher accuracy. For the feature extraction backbone, CSPDarknet-53 is used, which combines the Darknet-53 of the previous YOLOv3 with cross-stage partial (CSP) connections. Spatial pyramid pooling (SPP), which increases the receptive field, differentiates the essential features, and does not cause a speed reduction, is used as a neck over CSPDarknet-53. Moreover, instead of the feature pyramid network in YOLOv3, the path aggregation network (PANet) is used in YOLOv4. The original YOLOv3 network is used as the head. An AP of 43.5% was achieved Bochkovskiy et al. (2020); Yu & Zhang (2021); Mahto et al. (2020).

4. **Faster RCNN:** Faster R-CNN, proposed by Ren et al. (2015), is an extension of R-CNN and Fast R-CNN that improves object detection speed and efficiency. It comprises a region proposal network (RPN) and Fast R-CNN sub-network. RPN generates region proposals using anchor boxes from the backbone network's feature map, while Fast R-CNN classifies objects within proposals and refines bounding box coordinates using region of interest (RoI) pooling and fully connected layers. Backbone networks, such as VGG-16, ResNet-50, and ResNet-101, extract feature maps from input images. Faster R-CNN outperforms its predecessors in accuracy and speed. Faster R-CNN has a slower inference time compared to YOLO models due to the two-stage architecture. The model first generates region proposals and then classifies them, which increases computational complexity.

*2.2. Related Works*

Various approaches have been practised globally to address parking management and traffic-related issues. A work by Kher et al. (2020) introduces an AI-based smart parking management system, leveraging DL and Raspberry Pi for implementation. It offers real-time parking space availability updates, automatic billing, and flexibility for parking space owners. Another study focuses on traffic rule violation detection using DL, particularly YOLOv3, achieving high precision in identifying violations like helmet-less riders and triple-seat bikes Kathane et al. (2022). The work by Malireddi et al. (2019) presents an efficient pedestrian detection approach, employing fuzzy clustering and Histogram of Oriented Gradients, with a 92% classification accuracy. Moreover, the study by Agrawal et al. (2020) proposes an automatic traffic accident detection system that combines clustering, feature extraction with ResNet, and SVM classification, achieving an impressive 94.14% accuracy.

While these papers make valuable contributions to traffic and parking management, there is limited research addressing disability parking management systems. Although some studies have proposed DL-based ALPR systems, practical applications in real-world settings remain scarce. A work by Wang et al. (2021b) presents a DL-based Korean LPR system that achieved the highest recognition accuracy of 98.4% with the aid of 10,500 real images, 500,000 synthetic images, and extensive pre-processing. However, the study lacks discussion on the access badge or accessible parking management system, and the system appears to be complex. Another study by Usmankhujaev et al. (2020) developed a network that combines scene text recognition techniques with geometrical image transformation, achieving detection and recognition accuracies of 99.8% and 95.7%, respectively. They trained their model on approximately 50,000 CCTV images and created 60,000 synthetic images using OpenCV libraries. Furthermore, an end-to-end vehicle and LP detection model with a multi-branch attention neural network is proposed by Chen et al. (2019), where separate convolutional layers were developed for vehicle and LP recognition. In contrast Li et al. (2018) proposed a jointly trained network for simultaneous car LP detection and recognition without any intermediate processing, utilizing an end-to-end approach.

Likewise, the work of Gnanaprakash et al. (2021) proposes a YOLO-based tracking and number plate recognition system using roadside surveillance cameras. The proposed method comprises four steps: the conversion of video footage into images, the detection of cars within each frame, the recognition of number plates (NP), and character recognition of the NP. The Stanford car dataset was used for training the model, and LPs from Tamil Nadu, India, were used to evaluate the system performance. The system achieved 90% accuracy for character recognition using Gaussian filters to eliminate noise in the images and Canny's edge detection algorithm to detect edges. This method appears complex, and the version of YOLO used in their research remains undisclosed. Another study, Sahoo (2022) proposed a method for recognizing Indian vehicle license plates using the Prewitt filter technique to extract features of characters from the LP. Three classifiers, namely k-nearest neighbour, artificial neural network, and decision tree were utilized for classification. The proposed approach achieved a recognition accuracy of up to 98.10%.

Similarly, a layout-independent LP detection and recognition model was proposed by Seo & Kang (2022), utilizing the idea of CenterNet and attention-based recognition networks, tested on the Korean handicapped parking card dataset. Though this research is the only one found to incorporate the access badges provided by the Korean government, it does not provide any information regarding accessible parking area violations and solely focuses on LP detection. Again, in the work of Srividhya et al. (2022), a vehicle classification approach is presented that employs delta learning and machine learning algorithm to detect, segment and classify vehicles by eliminating their shadow counterparts. The proposed system is trained with various types of vehicles based on their appearance, colors, and build types. Furthermore, the paper outlines a method for recognizing the number plate using text correlation and edge dilation techniques. The authors reported different accuracies for NP detection as per frames per video, with the best accuracy of 92% achieved for 240 frames per video. In their research on license plate recognition, Islam et al. (2020) proposed a two-stage approach consisting of a detection stage and a recognition stage. In the detection stage, the input images were pre-processed to identify a RoI that contains the license plate. This RoI was then passed to the recognition stage, where the histogram of oriented gradients method was used to extract features from the license plate characters. These features are subsequently employed to train an artificial neural network for the identification of license plate characters. To extract the LP region, the authors utilized YOLOv2 object detection algorithm. Tu & Du (2022) proposed a multilevel hierarchical R-CNN-based method for detecting and recognizing vehicles and their LPs. Their method involves using a higher level of R-CNN to extract vehicles from the original images or video frames, followed by a lower-level R-CNN to detect the LP regions within the vehicle. Then, the detected LP are split into single numbers, and an even smaller R-CNN is used to recognize the individual numbers. The proposed approach achieved high accuracies of 98.5%, 97%, and 85% for vehicle detection, LP detection, and character recognition, respectively.

## 3. Methodology and Contribution

The methodology implemented in this research, which is straightforward and shown in figure 4, aimed to develop a system that could detect and recognize vehicles and access badges of people with disabilities in Korea. The following steps were undertaken to achieve this objective:

- **Data collection**: Images of Korean private cars and access badges were collected from various sources, including public parking lots, government websites, and online image databases.

- **Labelling:** The collected images were annotated for the YOLO (*.txt*) and PASCAL VOC (*.xml*) format for object detection. The free online tool, makesense.ai ., was used for labelling.

- **Data splitting and Training:** The annotated images were split into two sets: a training set and a validation set. The training set was used to train the deep learning models, while the validation set was used to evaluate the performance of the models during training. Transfer learning was employed to train the models on the labelled images. Specifically, faster RCNN model and three different YOLO models (YOLOv7, YOLOv5s, and YOLOv4) were fine-tuned and trained on the labelled images.

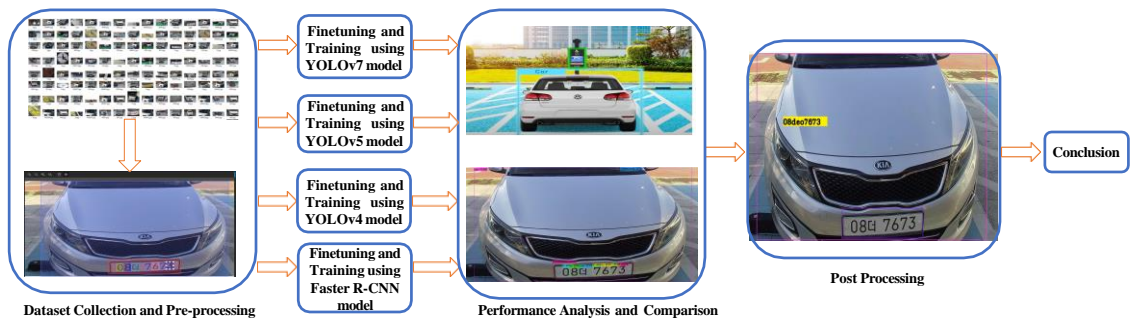

Figure 4: Overview of research methodology implemented in this research (Graphical Abstract).

- **Model Analysis:** The performance of the employed models was analysed using various evaluation metrics, including mean average precision (mAP), precision, and recall. These metrics were used to compare the performance of the models and select the best one for further analysis.

- **Post-Processing:** After the models were trained, post-processing was performed on the output of the models. Specifically, the vehicle and access badge numbers were sorted so that they could be displayed correctly in their original format.

- **Conclusion:** Finally, the conclusion was drawn based on the analysis of the results obtained from the models. The performance of the models was compared, and recommendations were made for future work to improve the accuracy and efficiency of the system.

Inspired by the widespread use of the YOLO series models and their efficiency for different OD tasks, we have implemented version 7 of this algorithm. Our research was prompted by the need for a proper system for the management of accessible parking in South Korea. Furthermore, none of the previous researchers incorporated access badges in their investigations. The novelty of this research lies in utilizing LP number, badges and card numbers for cross-validation with a central database containing all relevant information. This approach ensures a more reliable and accurate identification of vehicles with authorized access to accessible parking spaces. Additionally, we carefully curated our dataset to encompass various challenging conditions, including scenarios with low light, fog, and low-contrast situations. This comprehensive dataset empowers our model to exhibit robust performance across a broad spectrum of real-world scenarios. The main contributions of this research are:

- A single DL-based OD network is proposed for recognising the vehicle, LP, LP number, access badges, and badge number. The model works efficiently irrespective of the LP and badge size or type and performs well in all possible conditions.

- The dataset used in this research is a real-time scenario, i.e., collected from the different parking areas of South Korea, which improved the model testing performance. Thus, this model is proposed for South Korean vehicles and LPs only, but this approach can be used around the globe. We have implemented this model to another dataset as well, which we have mentioned in detail in Appendix B. The only thing needed is dataset collection and proper labelling, and the method can be implemented easily.

- No special pre-processing is required except for labelling the images, which is the must-do for any OD model. Simple mathematics-based post-processing is applied to recover the LP and card numbers in their original format.

- The cross-validation is done with the help of a central database containing the information regarding LP and access badges for checking the rights of a vehicle to use an accessible parking area to avoid the abuse of accessible parking space.

4. 4. Experimental Analysis

In this section, we present a comprehensive experimental analysis of the proposed *Shine* system for accessible parking management. Here, we provide an overview of the dataset utilized for training and validation, detail the DL models employed, and outline the performance metrics used for evaluation. Furthermore, we present the outcomes obtained from various models, including different versions of YOLO and the Faster R-CNN, and conduct a comparative analysis of their effectiveness in recognizing vehicles, LPs, and access badges. The experimental analysis serves to highlight the system's robustness and efficiency in the context of accessible parking space management, effectively addressing the issue of parking space abuse.

*4.1. Data Collection*

We have collected more than 30,000 images of cars, Korean license plates (*type a* to *type d*), and access badges–yellow, white, and brown cards, as shown in figure 1(b). The images were collected by clicking the vehicles and badges with the smartphones and downloading the freely available images. While clicking the photos, we first took the full image of a vehicle, displaying the car, LP, and cards (if available), then clicked another zoomed-in picture of LP only, and finally, the zoomed-in view of cards only. In this way, three images can be collected from a single vehicle. We visited the parking areas, disability welfare centres, shopping malls, hospitals, and so on to manage the images. We captured pictures in various conditions like low light, shiny, foggy, rainy, etc. The photographs are mainly collected from the parking area; thus, we believe these images best fit the real-time application. Usually, people are expected to place an access badge, as shown in figure 1(a); nevertheless, it does not always happen. The cards are found to be placed at any possible angle. Thus, we have used the data augmentation technique for cards, where we rotate the cards up to 60 degrees clockwise and anti-clockwise in an interval of 30 degrees. We had first used 360 degrees rotation; however, that generated errors because the digits like '6' and '9' created ambiguity, and the accuracy was dropped. Thus, we used 60 degrees rotation in either direction.

*4.2. Ground Truth Generation*

Next, we selected the images fit for our research. This was done manually. About 29,000 pictures were chosen for training the model and then labelled in YOLO format (*.txt* format). Labelling is the most crucial step in using the YOLO model. If the images are labelled carefully, half of the problem in OD is solved. There are some rules in labelling that the researcher should follow strictly to get better results. There are many open-source annotating tools found on the internet. We have used the *www.makesense.ai* website for labelling the objects inside the images. Makesense.ai is a free online tool for captioning photos. It uses a browser, so there is no cumbersome installation required, and it is ready to use as soon as the website is accessed. Again, it is accepted in any operating system. It is ideal for small computer vision deep learning projects, as it makes preparing datasets easier and faster. The labels created can be downloaded in many supported formats: VOC XML, YOLO, CSV, and JSON. Once the images were finished labelling, the pictures and labels were split into the training and validation sets at the ratio of 0.9:0.1, i.e., 90% training and 10% validation sets. The number of classes used in this research is 69, which includes the vehicle, LP type: electronic, green plate, white plates or the new ones with hologram, the characters used in the LP, digits used in LP and cards. The data distribution across each class is presented in figure 5. The names of all the classes used is shown in the x-axis and the number of the images each class occupies is shown in y-axis. The 'car' class contains the largest number of images, which is followed by the class 'sincarplate'. It comprises all white license plates, including new hologram plates. In contrast, the 'card' and 'gucard' classes have a relatively small number of items. This is likely because these types of cards are no longer in use and have been replaced by yellow, white, or brown cards. We chose to include a limited number of items in our dataset because they were challenging to obtain more of. This approach was intended to ensure that our model can still effectively detect and recognize these items in any context where they might appear.

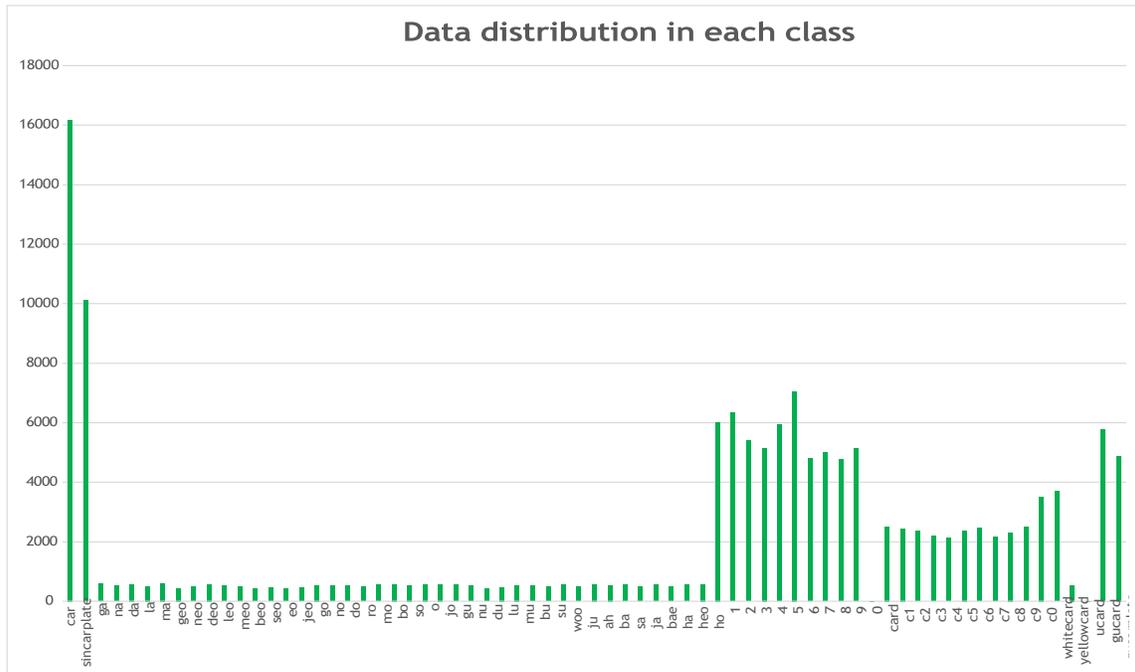

Figure 5: Data distribution across classes for training the employed models for this research. This chart presents the number of images available in each class used for training the models.

*4.3. Training and Testing YOLOv7 Model*

We have employed the TL approach to train our model in this study. Transfer learning allows models that have been previously trained on large image classification datasets to be adapted to specific tasks. Specifically, we have used the pre-trained YOLOv7 model, which was trained on MS COCO dataset from the scratch Wang et al. (2023). The pre-trained YOLOv7 model can detect 80 classes, including 7 vehicle categories: car, bicycle, motorcycle, bus, airplane, train, and truck. As these categories comprise 'car', our choice to employ the pre-trained YOLOv7 model for this project is both rational and justified. As mentioned earlier, the pre-trained YOLOv7 model was trained for 80 classes, thus, we need to finetune the model according to our requirement. The batch size of 32 and image size of 640×640 were used for training the YOLOv7 model. The number of classes was changed to 69, and the filters for last 3 layers were changed accordingly. Since the dataset contains both small and large objects, the respective changes in the configuration file were made: like changing the number of scales or output layers, number of strides, number of anchors and so on. The model was trained for 300 epochs in the PyTorch framework Paszke et al. (2019). The Nvidia GeForce RTX 3090 GPU was used for training the model. The initial learning rate of the model is 0.01, the momentum of the learning rate is 0.94, the optimizer adopts stochastic gradient descent.

We evaluated the performance of our trained model using classification metrics, which allowed us to assess its ability to accurately detect and identify vehicles, license plates, license plate numbers, and cars. To assess the performance of object detection models such as YOLO, the mAP is a commonly used metric. This metric compares the detected bounding box to the ground-truth bounding box, resulting in a score. A higher mAP score indicates that the model has made more accurate detections. To validate our results further, evaluation metrics of precision, recall, and F1-score are employed. Precision is the measure of the number of true positive predictions divided by the total number of positive predictions. Recall is the measure of the number of true positive predictions divided by the total number of actual positives in the dataset. The F1-score is a metric that combines precision and recall, it is the harmonic mean of precision and recall. The equations for precision, recall and F1- score are defined as follows:

$$Precision = \frac{TP}{TP + FP} \quad (1)$$

$$Recall = \frac{TP}{TP + FN} \quad (2)$$

$$F1-score = 2 \times \frac{Precision \times Recall}{Precision + Recall} \quad (3)$$

where, TP, FP and FN are true positive, false positive and false negative, respectively Rafique et al. (2023).

Using YOLOv7 model, the mAP@0.5 was found to be 92.16%. The precision and recall were 94.7% and 90.84%, respectively. According to the above formula, F1-score is calculated to be 92.73%. The mAP@0.5:0.95 was found to be 66.57%. The mAPs@0.5:0.95 means starting from the intersection over union (IoU) = 0.5 increases to IoU = 0.95 in 0.05 steps. In this case, the AP threshold would be calculated with ten different IoUs. An average is done to provide a single number that rewards detectors better at localisation. The IoU is calculated as:

$$IOU = \frac{DetectionResult \cap GroundTruth}{DetectionResult \cup GroundTruth} \quad (4)$$

Figure 6 shows the model performance graphs. The figure consists of four individual graphs, each displaying the progression of a specific metric over time (epochs): Precision, Recall, mAP@0.5, and mAP@0.5:0.95. In each graph, the x-axis represents the number of epochs, while the y-axis corresponds to the metric's value for that particular plot. The confusion matrix obtained using the YOLOv7 model is shown in figure 7. The results obtained for all other classes are promising except for the classes 'card' and 'gucard'. As mentioned in the earlier section, these two classes contained very few items. Thus, the classification accuracy is impacted. However, this does not impact the proposed system because 'card' and 'gucard' are not in practice anymore and are replaced by yellow, brown, and white cards. Moreover, the results obtained on testing in various environmental conditions are shown in Appendix A. We have also implemented the proposed model with another dataset. The details are in Appendix B.

*4.4. Comparison*

The model testing was conducted on an Nvidia GeForce 1060 GPU machine, comparing YOLOv5s, YOLOv4, and Faster R-CNN models. To ensure fairness, parameters such as batch size, image size, and epochs were kept consistent. With YOLOv5s it was easier to use with these parameters than with YOLOv4, as the latter relies on the Darknet framework instead of PyTorch, which is employed by both YOLOv7 and YOLOv5s. The Faster R-CNN model, which can be employed in both TensorFlow and PyTorch, is implemented it in PyTorch framework for this project.

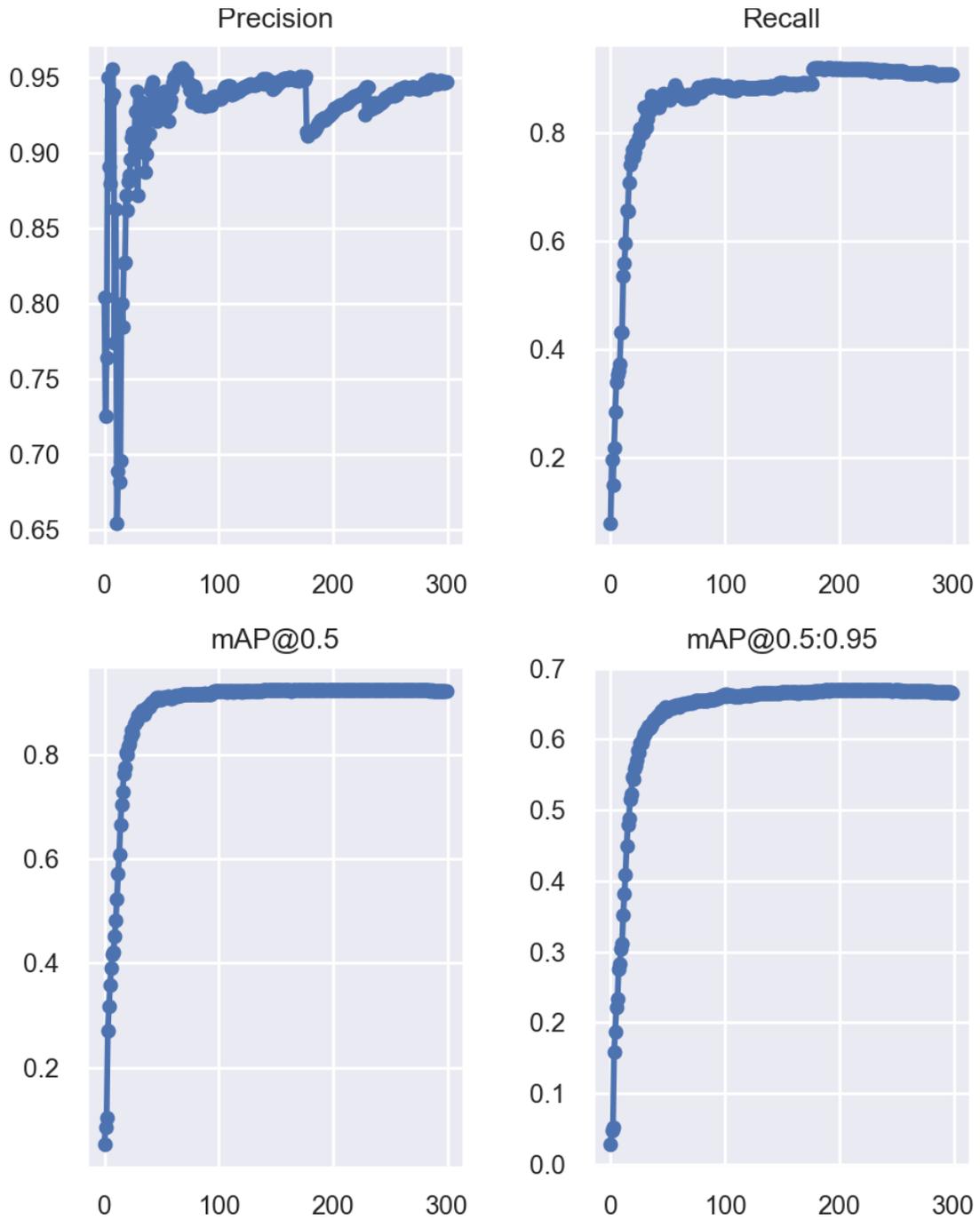

Figure 6: Model (YOLOv7) performance metrics over time (epochs). The figure comprises four separate graphs, each illustrating the evolution of a specific metric—Precision, Recall, mAP@0.5, and mAP@0.5:0.95—throughout the training process. The x-axis denotes the number of epochs, while the y-axis represents the corresponding metric value, offering a comprehensive understanding of the models' performance during training.

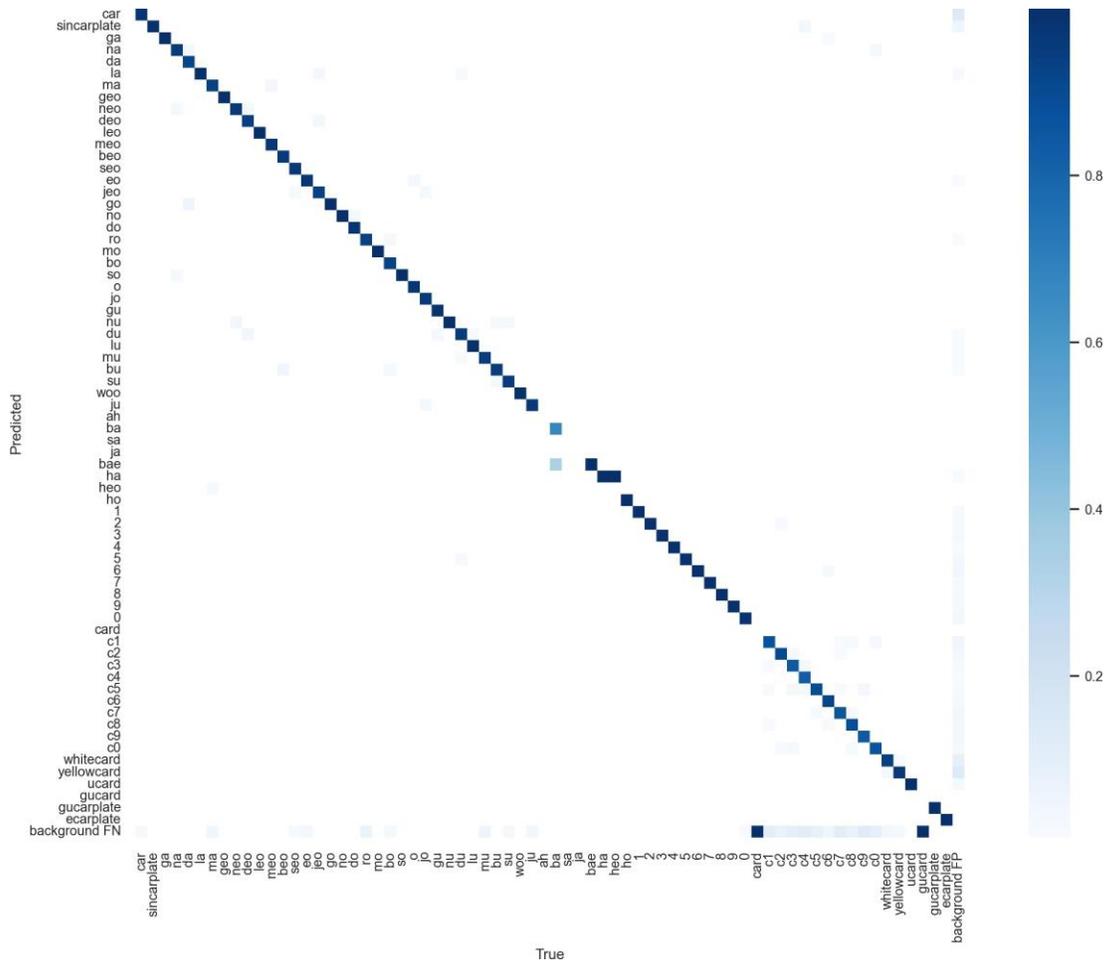

Figure 7: Confusion matrix of the proposed model (YOLOv7). This matrix visually represents the performance of the YOLOv7 model in classifying objects, highlighting the relationship between true and predicted classes.

*4.4.1. With YOLOv5s*

We trained our data to the YOLOv5s pre-trained model. The mAP@50 was found to be 87.7753. The precision and recall were 94.51 and 83.08 per cent, respectively. The mAP@0.5:0.95 was found to be 62.13%. Furthermore, the average inference time taken for detection from a picture and a video were found to be 12.63 and 13.10 milli seconds, respectively. The average (of three repetitions) total time taken for a video of 3 seconds was found to be 9.95 seconds.

*4.4.2. With YOLOv4*

Training the same dataset to the YOLOv4 model, the mAP@50 was found to be 83.17%. The precision and recall were 92% and 87%, respectively. Moreover, the average inference time taken for detection from a picture and a video were found to be 260.1 and 97.96 milli seconds, respectively. The average (of three repetitions) total time taken for a video of 3 seconds was found to be 18.32 seconds.

*4.4.3. With Faster R-CNN*

We also trained the same dataset to the Faster R-CNN model using Resnet-50 as a backbone. The mAP@50 was found to be 70.50%. The precision and recall were 69.94% and 72.9%, respectively. Moreover, the average inference time taken for detection from a picture and a video were found to be 292.33.1 and 171 milli seconds, respectively. The average (of three repetitions) total time taken for a video of 3 seconds was found to be 31.99 seconds. The evaluation of the performance of the models used is presented in figures 8 and 9. Figure 8 depicts a graphical representation of the performance metrics for each YOLO versions and Faster R-CNN model utilized in this study. Furthermore, figure 9 compares the inference and total time required by these models for detecting objects in both image and a three-second video. To ensure statistical validity, the performance evaluation was conducted on three repetitions for and image and video, with the average value recorded for each YOLO version. The outcomes of our analysis reveal that the YOLOv7 model demonstrates superior accuracy when compared to the other three models, as demonstrated by the results presented in figure 8. Nevertheless, YOLOv5 outperforms the other models in terms of computational speed, as illustrated in Figure 9. Given the prioritization of accuracy in this project and the acceptable speed of the YOLOv7 model, its adoption for this research is deemed appropriate and justified.

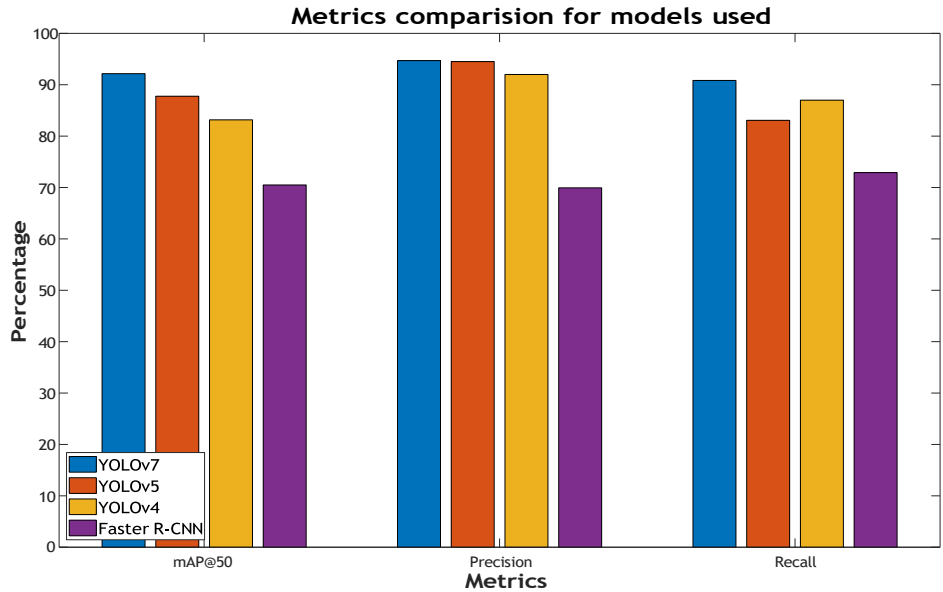

Figure 8: Performance metrics comparison: YOLOv7, YOLOv5s, YOLOv4 and Faster R-CNN.

*4.5. Post Processing*

This is the final step of this project. After the detection is done using the pre-trained YOLOv7 model, the detection results are post-processed. The detection result from the YOLOv7 model is processed to get the plate and card numbers in a proper orientation. Figure 10 shows the post-processing for the vehicle and card number. We have implemented simple mathematics for this part. As soon as the model detects the plate and card type–white, hologram, electric, or green plate, brown, yellow, or white card—-further processing is done, in which the bounding box coordinates are taken for reference to generate the plate number or card number as per their format (shown in Table 1). We have used the English translation for the Hangul characters in this step. The following section provides a concise overview of the *Shine* system, an ongoing project focused on object detection+. Given the importance of security and addressing potential concerns, we have intentionally limited the level of detail in our description. We invite readers to explore the promising potential of *Shine*, as we carefully balance transparency and safety.

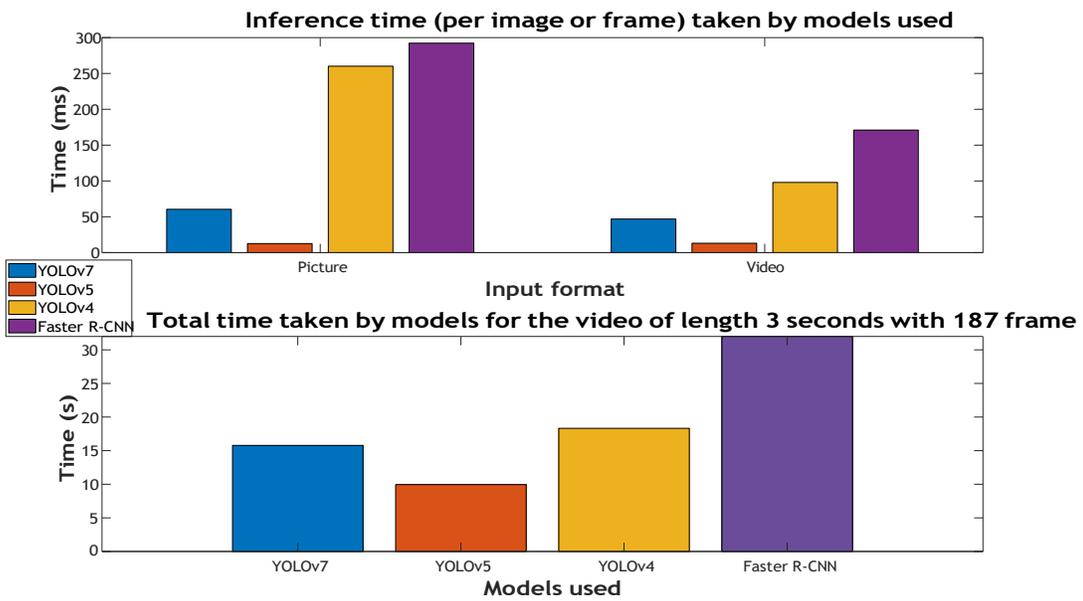

Figure 9: Inference and total time comparison for object detection models. This figure compares the time required by each of the YOLO versions and Faster R-CNN model for detecting objects in both image and a three-second video.

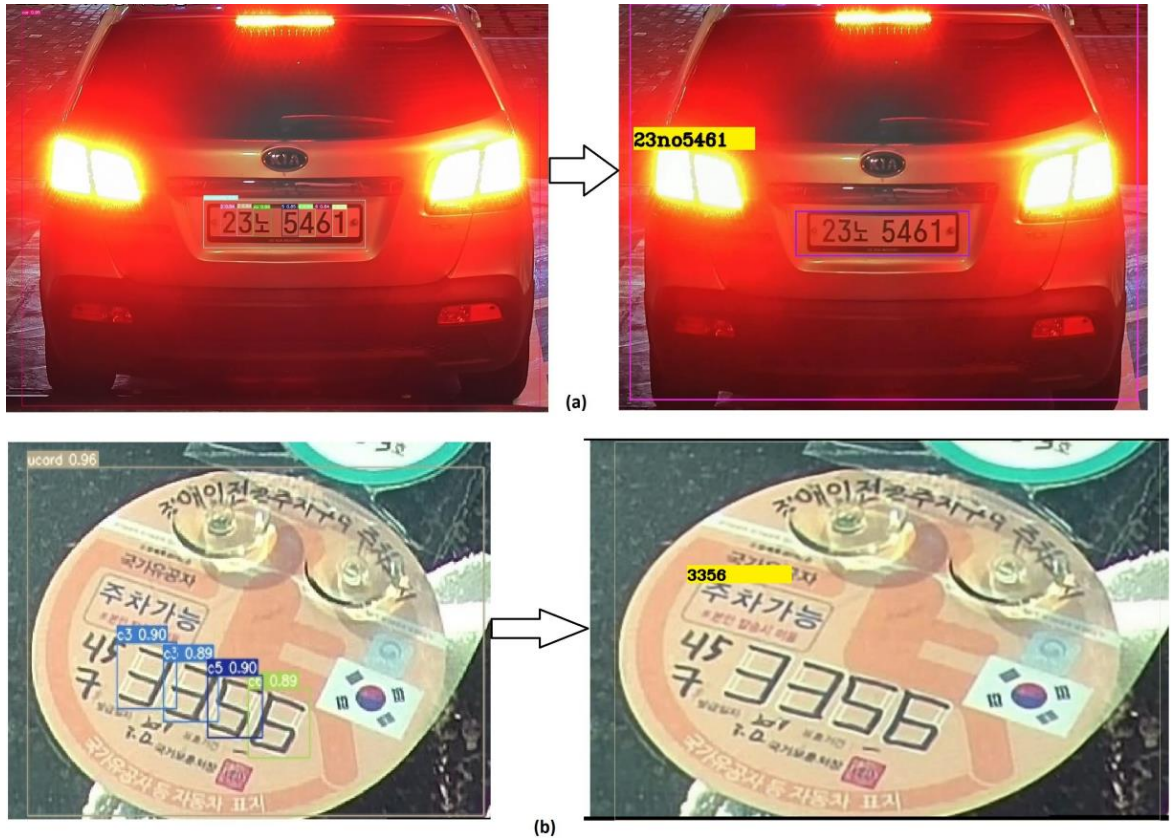

Figure 10: Post Processing of (a) Vehicle number, (b) Card number.

## 5. *Shine* – An Object Detection plus system

As mentioned in an earlier sections, *Shine* is an OD+ system. The '+' means it contains the object detection model plus other devices and algorithms that complete the accessible parking management system and, thus, mitigates the problem of accessible parking violations. Along with the OD model, this system combines a display monitor, a minicomputer with an Nvidia GPU, an IP camera, a speaker, LED strips, and a LED matrix board. Figure 11 provides information regarding the components, working, and algorithm of the *Shine* system; Figure 11(a) shows the contents of the *Shine* system, and figure 11(b) shows the working of the OD module used in *Shine*. It is explained in the earlier sections in detail. Figure 11(c) shows the further processing of the output from the OD model, which is used to verify the authenticity of the right to access accessible parking spaces.

When a vehicle approaches the accessible parking area, an IP camera mounted within our system, referred to as *Shine*, detects the vehicle. This camera then proceeds to identify the vehicle's LP number by zooming in on the plate. Additionally, it scans the entire front windshield for an access badge. Once the badge is located, the system zooms in to read its number. You can see this process depicted in Figure 11(b). Moreover, the information about the vehicle, LP, and badge is stored locally and transmitted to a central database. This central database contains details about the vehicle and the corresponding card number assigned to it. It cross-checks the link between the LP and the card and subsequently sends a response back to the *Shine* system. This step is depicted in Figure 11(c). If the vehicle is authorized to use the accessible parking area, a green LED blinks, and both the LED matrix board and the speaker provide a welcoming message. However, if the response from the central database is negative (indicating that the vehicle and card number are not linked), a red LED blinks, and a warning message is played through the speaker and displayed on the LED matrix board. This way, this approach serves to reduce the misuse of accessible parking spaces and ensures that fraudulent activities are met with appropriate consequences.

## 6. Challenges and Recommendations for ALPR

Accurate and fast LPR is one of the most important issues in license plate recognition systems. Due to the high frame rate of surveillance cameras, conventional LPR systems cannot be used in real-time. On the other hand, the presence of natural and artificial noise and light and weather conditions complicate the detection process for these systems Pirgazi et al. (2022). When using the DL and OD-based ALPR systems as well, the researchers encounter various challenges. Some of the challenges and probable recommendations are listed below:

- The biggest challenge is regarding the data. DL-based approaches need an enormous amount of data. Also, data from various conditions are required for real-time applications, which is a challenging task. Thus, researchers are advised to collect considerable data for implementing DL-based OD algorithms.

- The lack of availability of labelled datasets is another challenge. The researchers need to label the data manually.

- Annotation is a crucial part of the DL-based OD model. A proper labelling tool and procedure must be followed; otherwise, an OD model will not be very efficient. The improper labelling of the objects results in inaccurate localisation during prediction. Different easy and efficient annotating software and websites are available. The researchers should thoroughly research the tools and annotation tutorial before labelling the images.

- Training DL models is very computationally intensive and costly. The most advanced models might take several weeks to train on a high-performance GPU machine. In addition, it requires a large amount of memory to train the models Neupane & Seok (2020c).

- There is variability in the occupancy of objects in an image, with objects occupying most pixels, i.e., more than 80%, and sometimes occupying very few pixels, such as 10% or less. The researchers should use the respective configuration file for the respective object occupancy. The researchers must finetune the model configuration if the OD task needs objects occupying varying pixels. The authors of YOLOv4 have mentioned this issue in their GitHub repository.

- For detecting smaller objects more accurately, it is suggested to increase the image size while training.

- Many object detectors are trained on inputs of a particular resolution. The performance of these detectors is generally lower for a difference in resolution inputs. The researchers are recommended to set the exact training and testing image resolution for better accuracy Diwan et al. (2023).

- One common issue with OD models is the detection of unwanted objects, resulting in false negative detections. To mitigate this issue, it is advisable to incorporate images with non-labelled objects that researchers do not want the model to detect. Such images should be accompanied by an empty '.txt' file, which indicates the absence of any bounding box. This helps the OD model to learn from negative samples and reduce false negative detections [6].

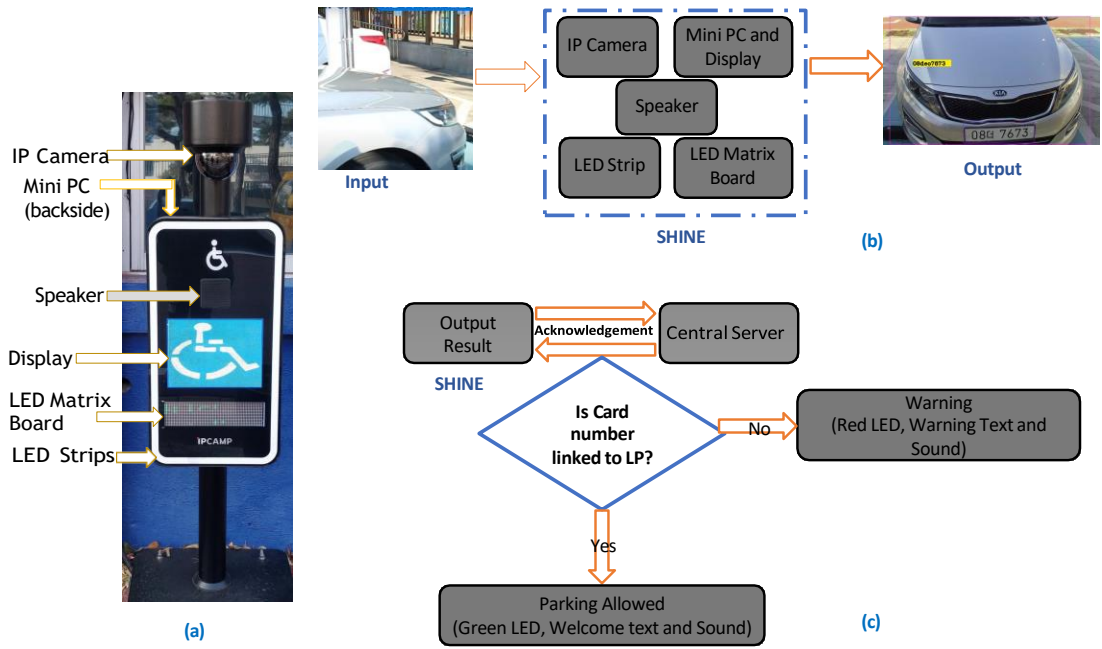

Figure 11: Components and workflow of *Shine* system: (a) shows the system's contents, (b) object detection module's operation, (c) demonstration of the post-processing of the OD model's output to verify accessible parking rights.

## 7. Discussion and Conclusion

DL-based algorithms have gained enormous attention in almost every field in this present age of artificial intelligence. With the growing concept of parking 4.0, many DL-based approaches are used in parking management systems. However, accessible parking management has a lot of gaps. Thus, to overcome the challenges and solve the problem of accessible parking space violations in South Korea, we have proposed a DL-based accessible parking management system, '*Shine*'.

*Shine* is a DL-based parking management solution designed for addressing the prevalent issue of parking related issues in accessible parking area in South Korea. We have employed the YOLOv7 object detection model for effectively detecting vehicles, access badges, and license plates. A comparison with YOLOv5s, YOLOv4, and Faster R-CNN models demonstrated the justifiable use of YOLOv7 in this context. Furthermore, the proposed system employs the real-time authentication by cross-validating with a central server to ensure the rights of an authorized vehicle to use the accessible parking space. The *Shine* system consists of a display monitor, a minicomputer with Nvidia GPU, IP camera, speakers, LED strips, LED matrix board and many other components, providing a comprehensive solution for managing accessible parking. Also, its adaptability allows it to be used in different countries by retraining the object recognition model to recognise the respective LPs, access badges. Despite its effectiveness, the *Shine* system has its limitations. It is currently incapable of providing information regarding the availability of accessible parking spaces for the respective vehicles. Addressing this limitation in the future could further improve the efficiency of the system in managing parking problems in urban environments.

In summary, the *Shine* system presents a promising solution for addressing the challenges associated with accessible parking management, employing DL algorithms. By incorporating the YOLOv7 object detection model and other essential components, our proposed system has proven its effective-ness in mitigating violations and misuse of accessible parking spaces in South Korea. Furthermore, its adaptability opens doors for potential implementation in other countries. Our future works will focus on enhancing the system's capabilities by providing real-time information about the availability of accessible parking spaces. We believe that this system makes a significant contribution to improving parking management in urban settings, making it more efficient and effective.

---

[6]https://github.com/AlexeyAB/darknethow-to-improve-object-detection

**Acknowledgements**

This research is financially supported by Changwon National University in 2023-2024. The authors express their sincere gratitude to all the employees of IP Camp for their help throughout the research journey.

## Appendix A: Supplement Results of the Proposed YOLOv7 Model using Korean LP and Card Dataset
This section contains the detection result of the proposed system using the South Korean LP and access badge dataset.

| YOLOv7 Detection | Post Processing | Number |
|---|---|---|
| 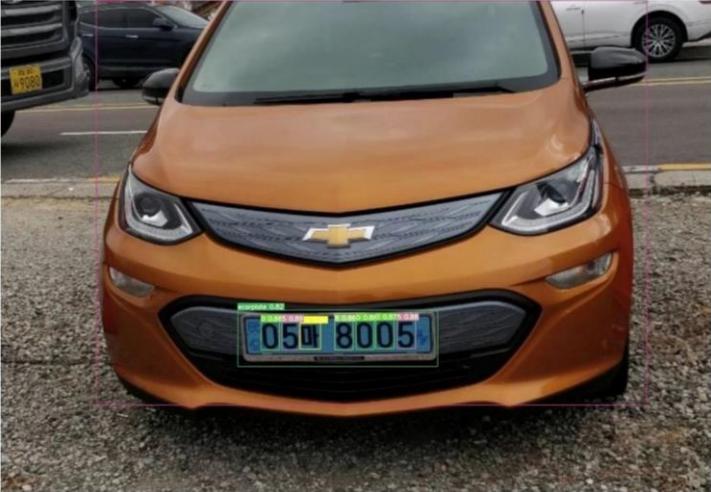 | 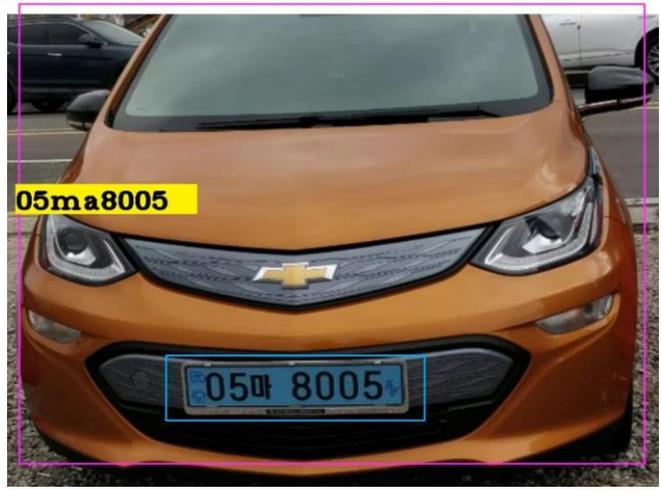 | 05 마 8005 |
| 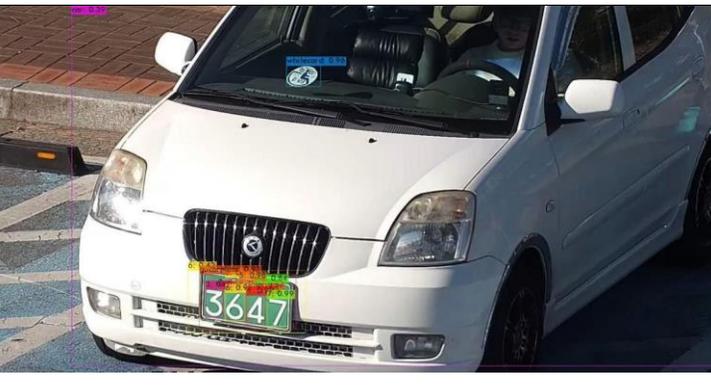 | 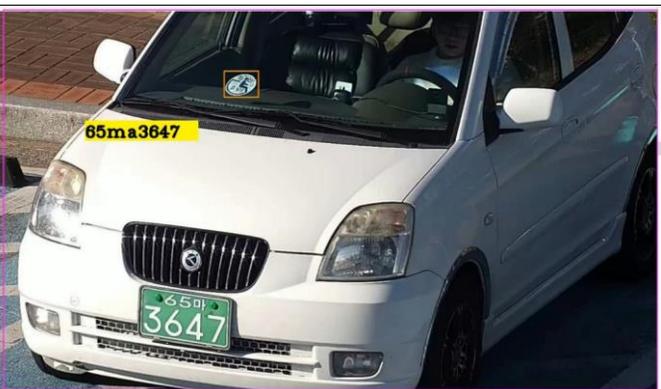 | 65 마 3647 |
| 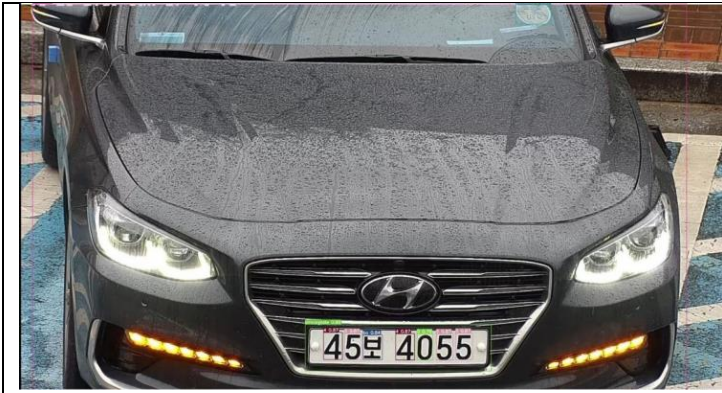 | 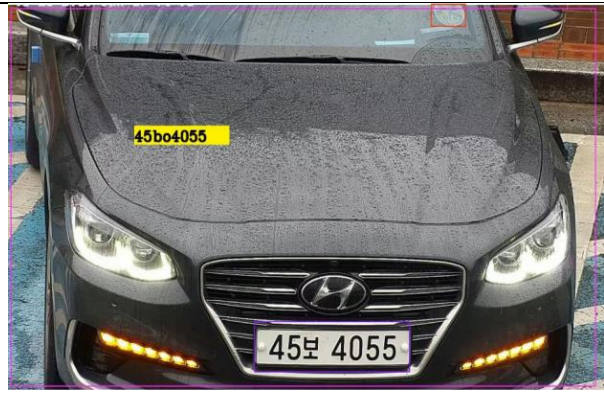 | 45 보 4055 |
| 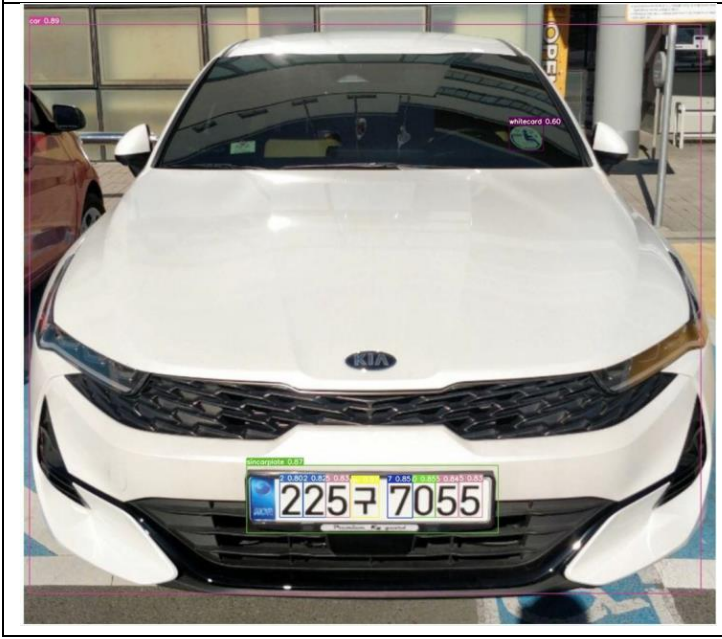 | 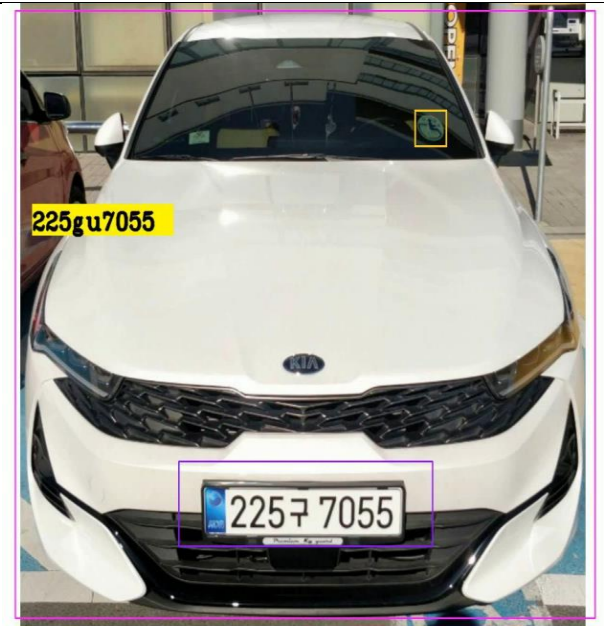 | 225 구 7055 |
| 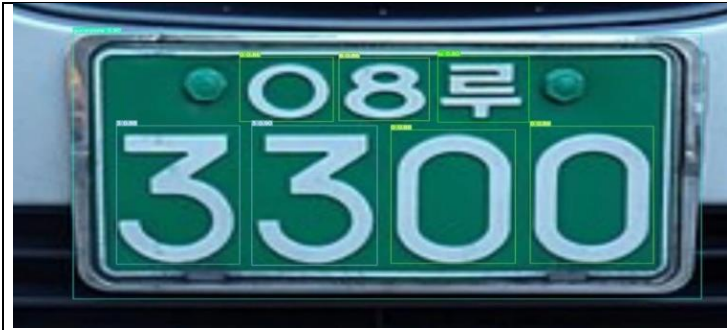 | 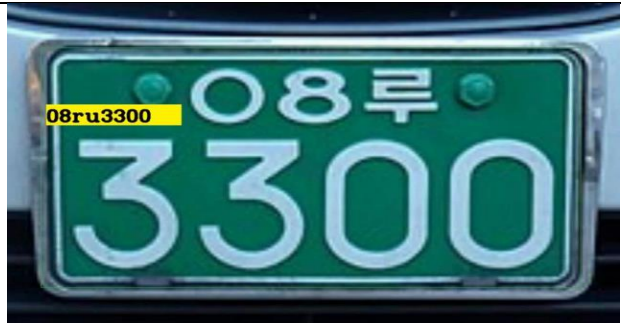 | 08 루 3300 |
| 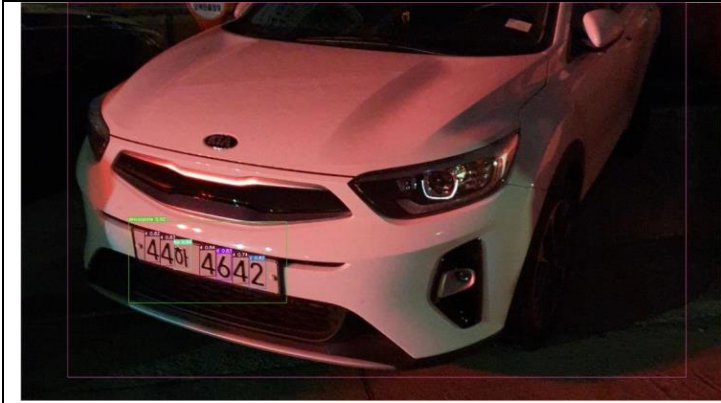 | 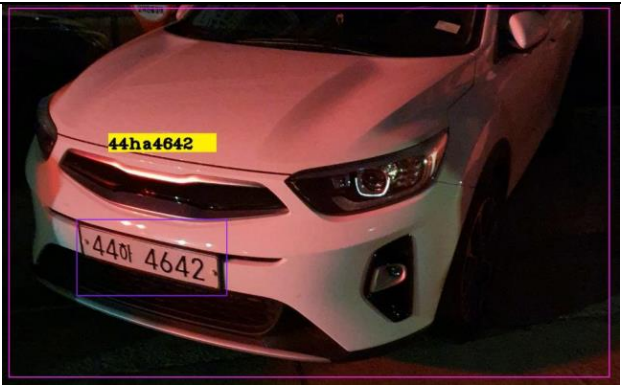 | 44 하 4642 |
| 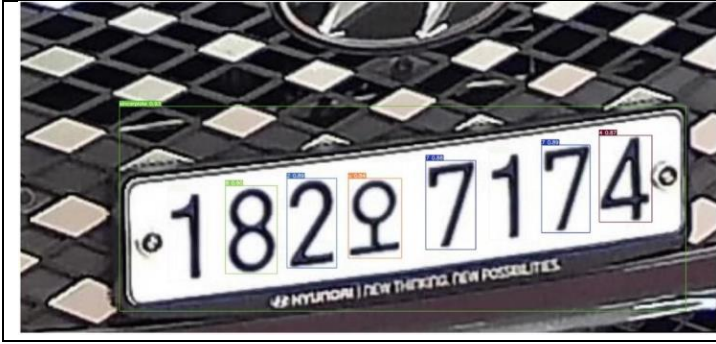 | 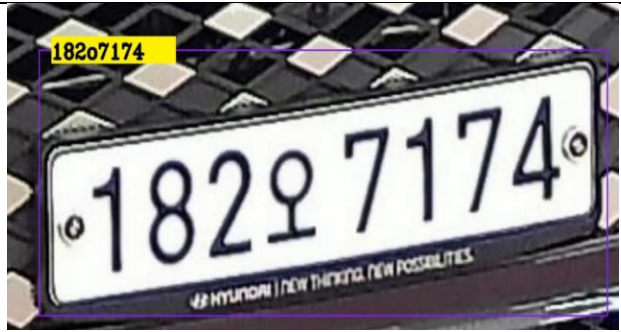 | 182 오 7174 |

| 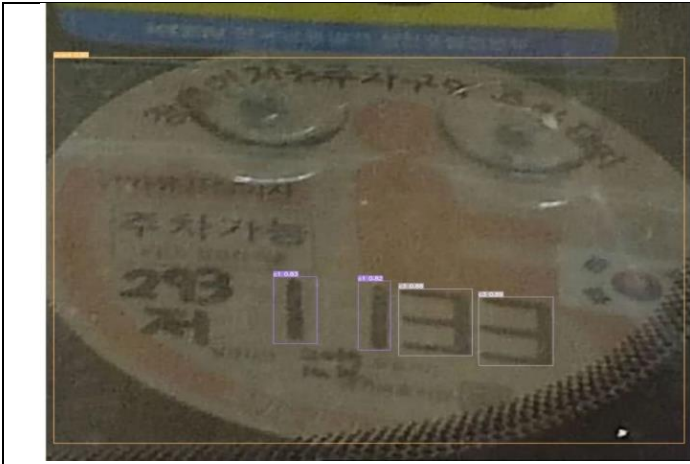 | 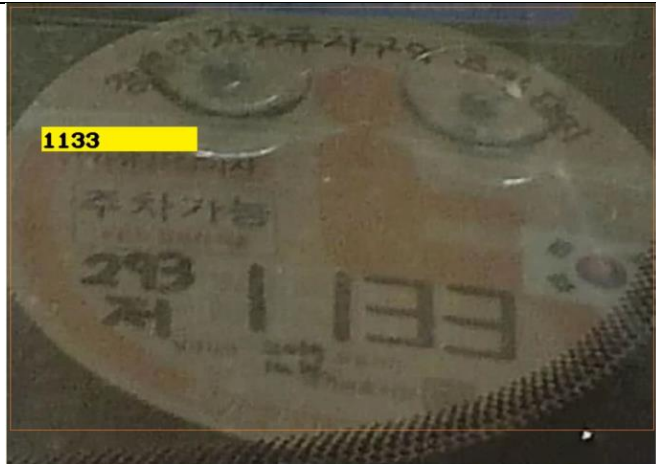 | 1133 |
| 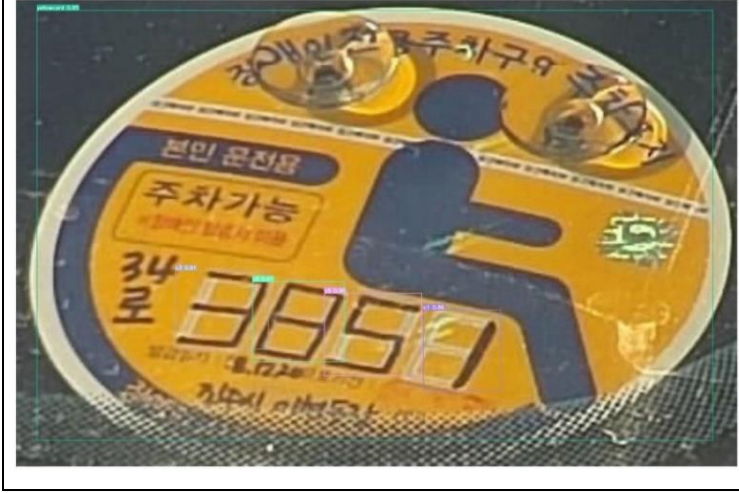 | 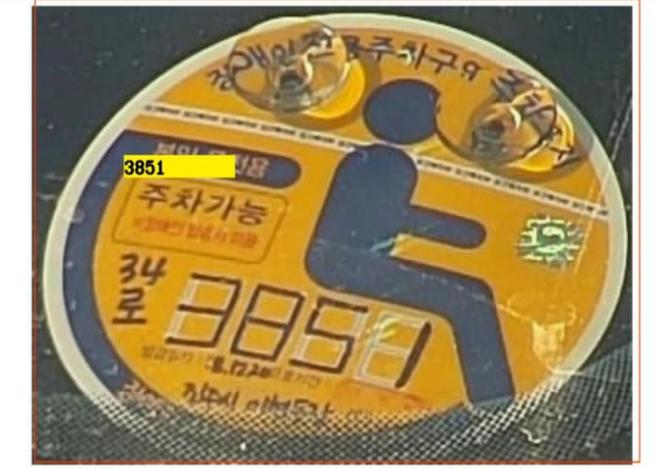 | 3851 |
| 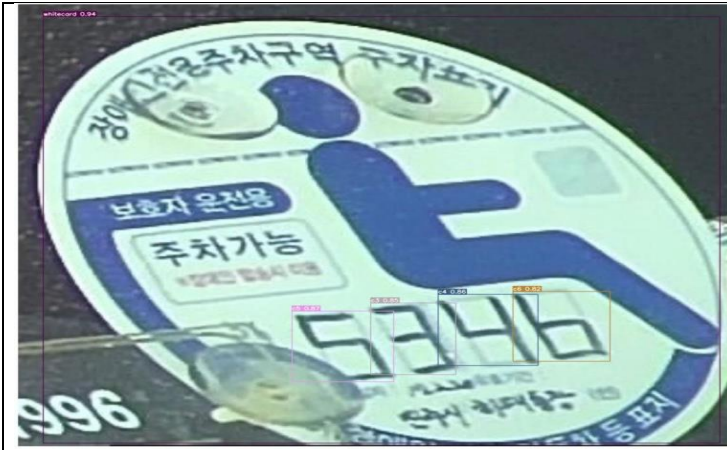 | 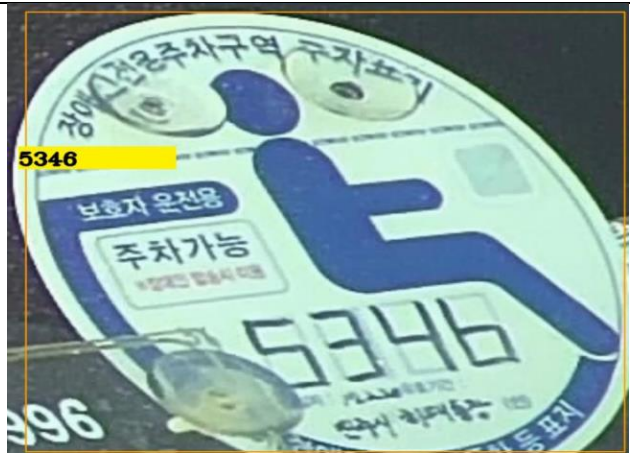 | 5346 |
| 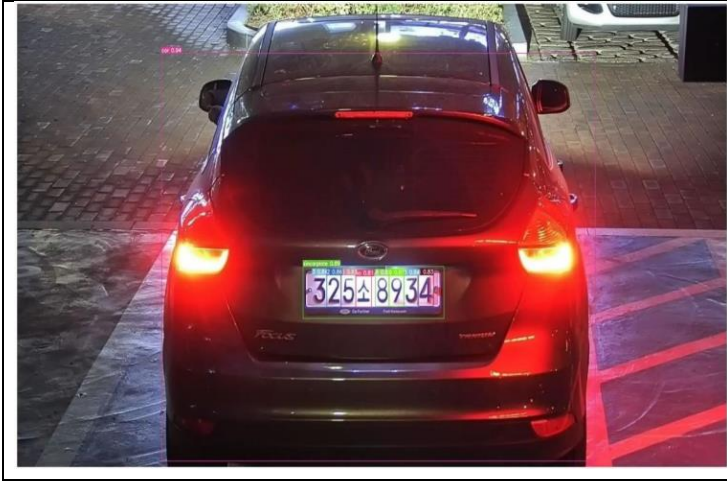 | 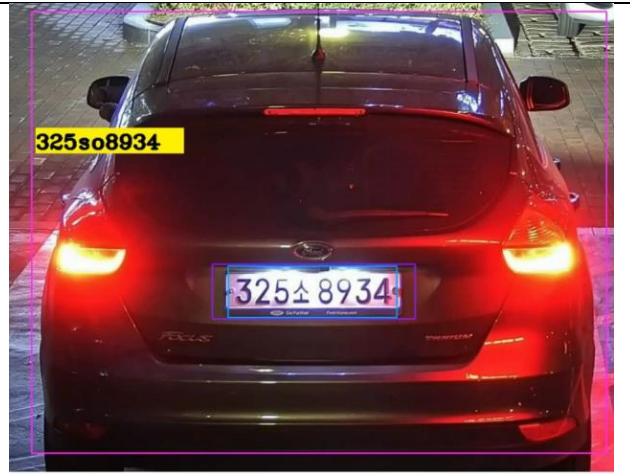 | 325 소 8934 |
| 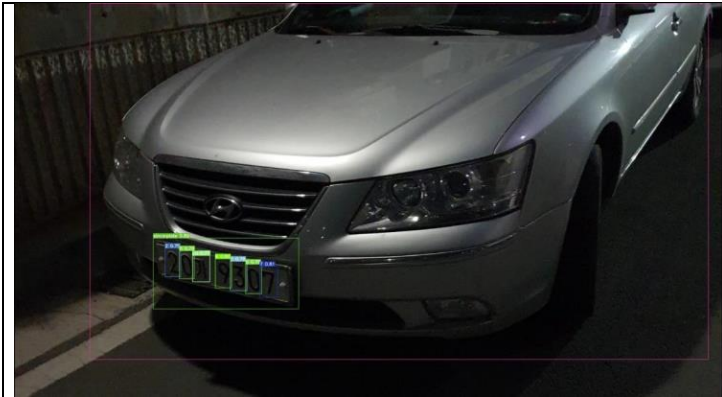 | 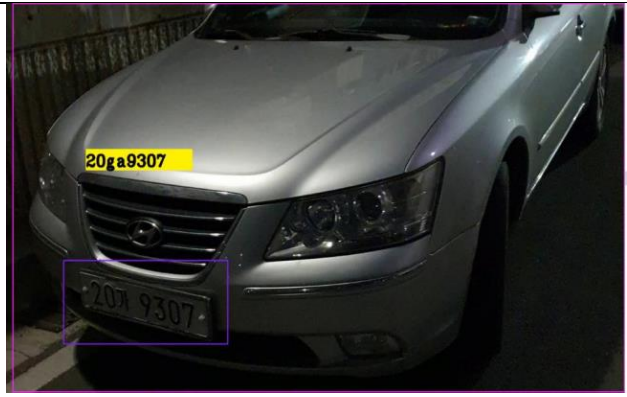 | 20 가 9307 |
| 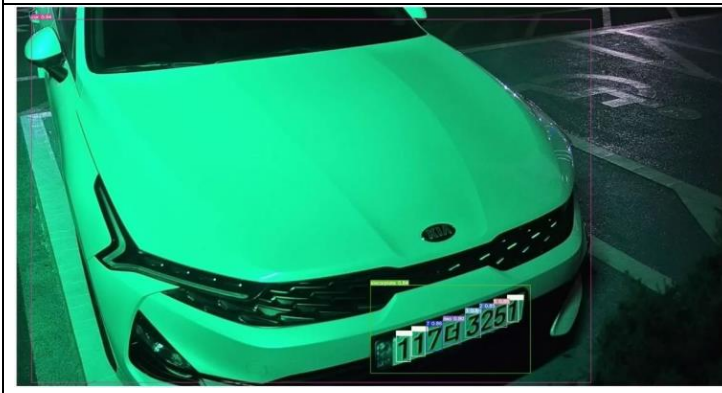 | 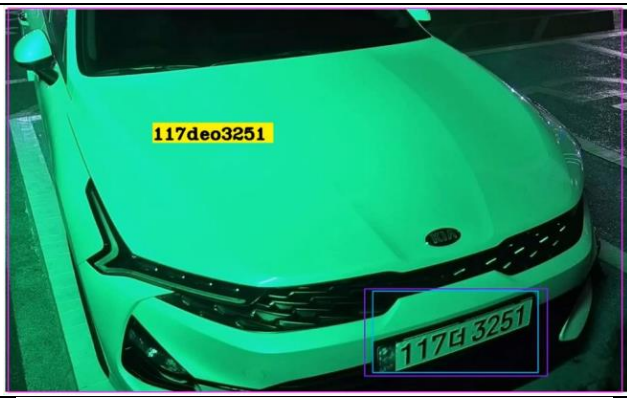 | 117 더 3251 |
| 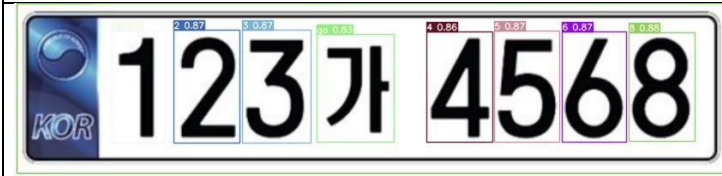 | 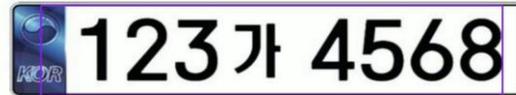 | 123 가 4568 |

**Appendix B: Evaluation of the YOLOv7 Model on an additional LP Dataset**

Appendix B presents an experiment in which the proposed YOLOv7 model is used on another LP dataset comprised of English alphabet characters and digits. Acquiring the substantial LP dataset is challenging due to privacy and security issues. Thus, we collected only about 500 images, further split into training and validation sets at the ratio of 0.9:0.1. The images were labelled in YOLO format (*.txt* format) using the *makesense.ai* online tool. The number of classes was set to 36, covering characters A-Z and digits 0-9. However, exceptions were made for I and O, as they are often replaced by 1 and 0 in many number plates, respectively. The model was trained using a similar setup environment as employed for the Korean license plate and cards dataset, ensuring consistency in the training conditions. Despite the relatively small dataset, the model achieved satisfactory results on the validation set with a precision of 0.827, recall of 0.812 and mAP@0.5 of 0.865. The result is relatively lower than those obtained for the Korean LP and access badge dataset, which can be attributed to the smaller number of training samples in this dataset. Nevertheless, this experiment demonstrates the robustness of the model and its potential application to number plate recognition using different character sets. The following figures show the model performance metrics over time (epochs).

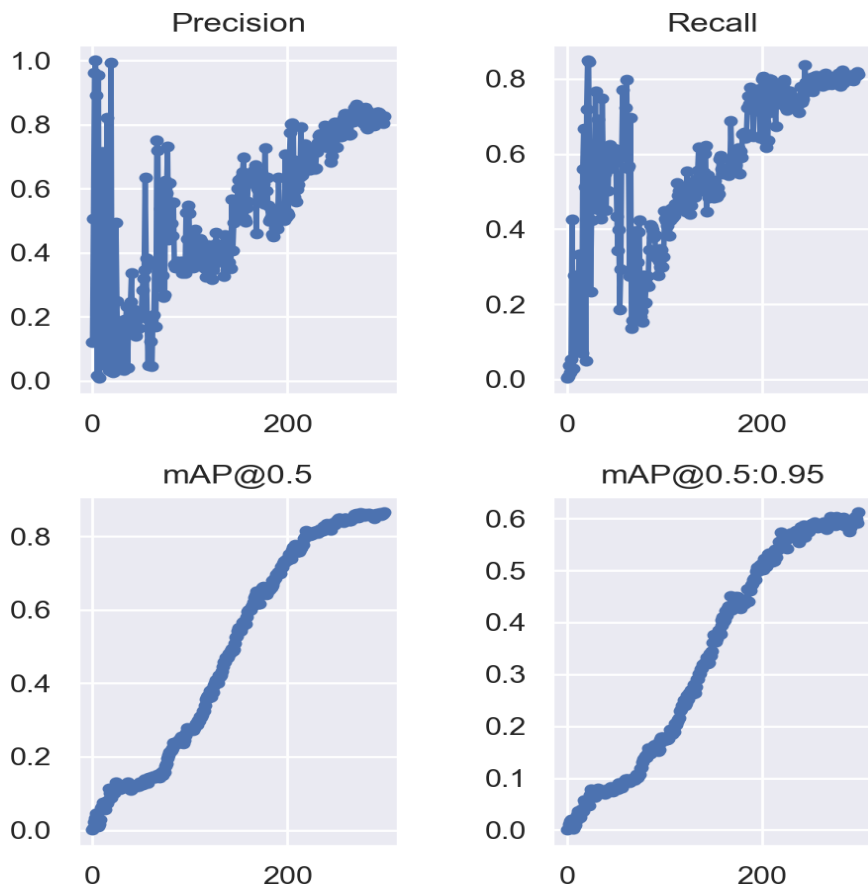

Figure .12: Model (YOLOv7) performance metrics over time (epochs). The figure comprises four separate graphs, each illustrating the evolution of a specific metric—Precision, Recall, mAP@0.5, and mAP@0.5:0.95—throughout the training process. The x-axis denotes the number of epochs, while the y-axis represents the corresponding metric value, offering a comprehensive understanding of the models' performance during training.

Figure .13: Confusion matrix of the proposed model (YOLOv7). This matrix visually represents the performance of the YOLOv7 model in classifying objects, highlighting the relationship between true and predicted classes.

Wang, C.-Y., Bochkovskiy, A., & Liao, H.-Y. M. (2023). Yolov7: Trainable bag-of-freebies sets new state-of-the-art for real-time object detectors. In *Proceedings of the IEEE/CVF Conference on Computer Vision and Pattern Recognition* (pp. 7464–7475).

Wang, C.-Y., Yeh, I.-H., & Liao, H.-Y. M. (2021a). You only learn one representation: Unified network for multiple tasks. *arXiv preprint arXiv:2105.04206*, .

Wang, H., Li, Y., Dang, L.-M., & Moon, H. (2021b). Robust korean license plate recognition based on deep neural networks. *Sensors*, *21*, 4140.

Yang, F., Zhang, X., & Liu, B. (2022). Video object tracking based on yolov7 and deepsort. *arXiv preprint arXiv:2207.12202*, .

Yu, J., & Zhang, W. (2021). Face mask wearing detection algorithm based on improved yolo-v4. *Sensors*, *21*, 3263.

Zhang, Y., Wang, C., Wang, X., Zeng, W., & Liu, W. (2021). Fairmot: On the fairness of detection and re-identification in multiple object tracking. *International Journal of Computer Vision*, *129*, 3069–3087.

Zhao, Z.-Q., Zheng, P., Xu, S.-t., & Wu, X. (2019). Object detection with deep learning: A review. *IEEE transactions on neural networks and learning systems*, *30*, 3212–3232.47